\documentclass{article}

\usepackage{microtype}
\usepackage{graphicx}
\usepackage{subfigure}
\usepackage{booktabs} 

\usepackage[hyphens]{url}
\usepackage[breaklinks=true]{hyperref}

\usepackage[accepted]{icml2023}

\usepackage{amsmath}
\usepackage{amssymb}
\usepackage{mathtools}
\usepackage{amsthm}
\usepackage{bbm}

\usepackage[capitalize,noabbrev]{cleveref}

\theoremstyle{plain}

\theoremstyle{definition}

\theoremstyle{remark}

\usepackage[textsize=tiny]{todonotes}

\begin{document}

\twocolumn[
\icmltitle{Are Good Explainers Secretly Human-in-the-Loop Active Learners?}



\icmlsetsymbol{equal}{*}

\begin{icmlauthorlist}
\icmlauthor{Emma Thuong Nguyen}{equal,comp}
\icmlauthor{Abhishek Ghose}{equal,comp}
\end{icmlauthorlist}

\icmlaffiliation{comp}{[24]7.ai, California, USA}

\icmlcorrespondingauthor{Emma Thuong  Nguyen}{emma.nguyen@247.ai}
\icmlcorrespondingauthor{Abhishek Ghose}{abhishek.ghose@247.ai}

\icmlkeywords{Machine Learning, Explainable AI, Active Learning, ICML}

\vskip 0.3in
]



\printAffiliationsAndNotice{\icmlEqualContribution} 

\begin{abstract}
Explainable AI (XAI) techniques have become popular for multiple use-cases in the past few years. Here we consider its use in studying model predictions to gather additional training data. We argue that this is equivalent to Active Learning, where the query strategy involves a human-in-the-loop. We provide a mathematical approximation for the role of the human, and present a general formalization of the end-to-end workflow. This enables us to rigorously compare this use with standard Active Learning algorithms, while allowing for extensions to the workflow. An added benefit is that their utility can be assessed via simulation instead of conducting expensive user-studies. We also present some initial promising results.
\end{abstract}

\section{Introduction}
\label{sec:intro}

\begin{figure*}[ht]
\vskip 0.2in
\begin{center}
\centerline{\includegraphics[width=2\columnwidth]{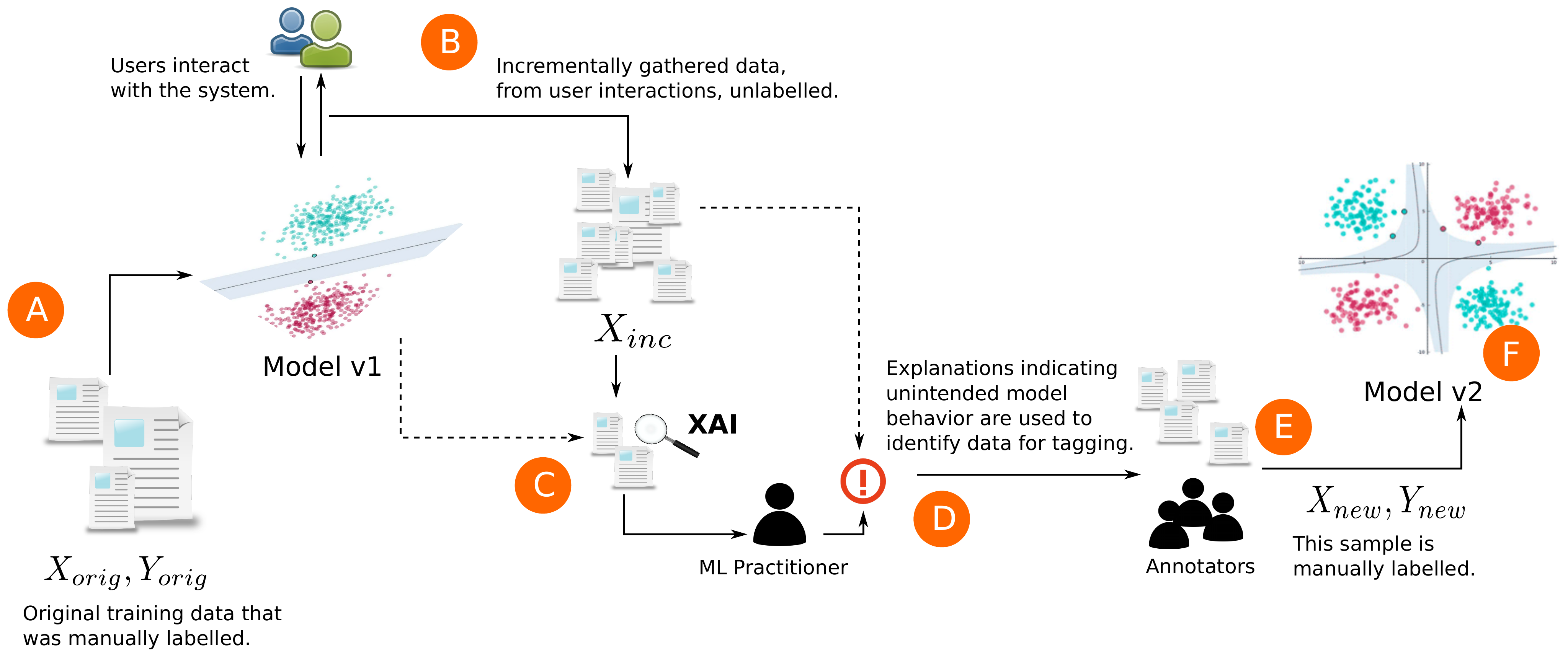}}
\caption{Workflow representing use of explanations to identify data to retrain a model. This shows one iteration of such a workflow, where we start with \emph{Model v1} and create a more accurate model \emph{Model v2}, based on sampling new training instances from a data pool, $X_{inc}$. Please see Section \ref{sec:intro} for details. As we shown in Equation \ref{eqn:one_eqn} in Section \ref{sec:formulation}, many of these steps may be distilled into a single mathematical expression.}
\label{fig:workflow}
\end{center}
\vskip -0.2in
\end{figure*}

Keeping in pace with the popularity of Machine Learning (ML), the past few years has seen a surge in the desire to understand how a model makes decisions. In certain domains, such as healthcare and law enforcement, such transparency is critical in acquiring the trust of its users. In others, it serves as a way to understand potential shortcomings of a system, e.g., if a system has overfit the data. This has led to accelerated research in the area of \emph{Explainable AI (XAI)}, which studies explaining of predictions from a given model, e.g. \emph{LIME} \citep{lime}, \emph{SHAP} \citep{shap}, \emph{DeepLIFT} \citep{deeplift}.


Here we consider the latter use of XAI: that of improving an ML classifier. We specifically look at the use of explanations to identify data that we deem ``interesting'' in some way, that is then used to further train our model. As an example, consider the workflow shown in Figure \ref{fig:workflow}, describing the following sequence of events:
\begin{enumerate}
    \item An ML practitioner trains a supervised classifier on an initial labeled dataset - $(X_{orig}, Y_{orig})$ - and deploys it (in Figure \ref{fig:workflow}, this is shown by \texttt{A}). The model is referred to as \emph{Model v1}.
    \item Users of this system interact with it, and in the process, incrementally generate (unlabeled) data, $X_{inc}$. Shown by \texttt{B} in Figure \ref{fig:workflow}.
    \item The ML practitioner periodically inspects the system for correctness. She samples from $X_{inc}$ and uses an explainer to review the model's decision process. Shown by \texttt{C} in Figure \ref{fig:workflow} - note that the model is required as an input to the explainer.
    
    Some explanations might indicate unintended behavior of the model. For example, both these reviews may be classified as positive, where the explainer has underlined the words that most influenced the classifier's decision:
    \begin{itemize}
        \item I \underline{love} the \underline{food} \underline{here}!
        \item I gave them a 1-star rating - that's how much I \underline{like} the \underline{food} \underline{here}.
    \end{itemize}
    Of course, the second review is sarcastic, and should be identified as negative.
    \item The ML practitioner decides to sample more such  examples from $X_{inc}$ (shown by \texttt{D} in Figure \ref{fig:workflow}), and then has them labeled by human annotators (shown by \texttt{E}). This new dataset\footnote{This new dataset may be seen to contain only the newly identified instances if the model may be incrementally trained, or a combination of the original data and the new instances, if the model needs to be trained from scratch. We will adopt the latter convention here since its universal, i.e., not all models support incremental training.} is denoted by $(X_{new}, Y_{new})$, and is used to further train the model to obtain \emph{Model v2} (shown by \texttt{F}). 
    
\end{enumerate}
This process is repeated multiple times to generate improved versions of models. Figure \ref{fig:workflow} shows one such iteration.
While this process seems intuitively appealing, we make rigorous the following aspects:
\begin{enumerate}
    \item First, we claim that this process essentially is \emph{Active Learning} \citep{settles2009active} that involves a human-in-the-loop (Section \ref{sec:activelearning}).
    \item Then, we mathematically formulate the workflow, which makes it convenient to (a) quantify its utility, and (b) compare with other AL techniques such as \emph{margin-based sampling} \citep{margin10.1007/3-540-44816-0_31} (Section \ref{sec:formulation}).
    
    An added benefit is such workflows may be evaluated via simulation bypassing the need for conducting time-consuming or expensive user studies.
\end{enumerate}

Our primary interest is text classification, but much of the discussion here applies to other forms of data as well.

\textbf{Related Work}: Based on a thorough search of the relevant literature, we believe this is the first study that casts human-in-the-loop data selection based on explanations as an AL query strategy. Other intersections of AL and XAI have been studied however, e.g., \citet{alden10.1145/3459637.3482080}  uses certain properties of local explanations to determine instance-informativeness for querying, \citet{10.1145/3432934} studies the benefits of annotating queried instances with explanations to obtain rich feedback from a human labeler.

\section{XAI-based Data Selection is Active Learning}
\label{sec:activelearning}
In many situations, while abundant unlabeled data is available, labeled data may be hard to procure, e.g., when manual annotation by experts is required. In such cases, one needs to explicitly account for the \emph{label acquisition cost}. The \emph{Active Learning (AL)} family of techniques solves for this problem in the following way:
\begin{enumerate}
    \item An initial model is built by acquiring labels for a small batch of data. This batch may be randomly selected from the unlabeled pool of data.
    \item The model is then iteratively improved by \emph{strategically} selecting data (from the unlabeled pool) to be annotated. This data is then used to further train the model. Such a strategy or \emph{query strategy}\footnote{So called because it is used to query instances from the unlabeled pool.} picks instances that have the greatest influence on the model's accuracy. Some popular query strategies are \emph{entropy sampling} and \emph{maximum margin sampling}.
\end{enumerate}

Informally, the query strategy is a mechanism to identify maximally useful instances given (a) the current model, and (b) an unlabeled pool of data. Referring to Figure \ref{fig:workflow}, we observe the following components effectively form a query strategy:
\begin{itemize}
    \item The explainer used to detect surprising patterns in model predictions (shown by \texttt{C} in Figure \ref{fig:workflow}).
    \item The process of using the explanations to solicit further instances from the unlabeled pool $X_{inc}$ (\texttt{D} in Figure \ref{fig:workflow}).
\end{itemize}

Specifically, this is the \emph{batch} AL setting, where batches of data are iteratively identified, labeled and used to train the model, e.g., \emph{BatchBALD} \citep{NEURIPS2019_95323660}. AL may be used in various other settings as well, such as \emph{stream-based} - see \citet{settles2009active} for an overview.

\section{Mathematical Formulation}
\label{sec:formulation}
How do we compare this form of AL with standard AL techniques? Clearly, a challenge is that because there is a human-in-the-loop - the ML practitioner - this workflow needs to be tested with expensive or time consuming user-studies. In this section, we try to eliminate this roadblock by (1) providing a reasonable approximation for  the task of the ML practitioner, and (2) offering a concise representation for the overall workflow. This makes it possible to efficiently simulate the workflow from Figure \ref{fig:workflow}.

We introduce some notation first:
\begin{enumerate}
    \item We will denote the number of instances in the collection $X_a$ by $N_a$. We will also assume that our data resides in $d$ dimensions, i.e., $X_{orig} \in \mathbb{R}^{N_{orig} \times d}$, $X_{inc} \in \mathbb{R}^{N_{inc} \times d}$ and $X_{new} \in \mathbb{R}^{N_{new} \times d}$.
    
    \item We will assume explanations are produced in $d'$ dimensions. The case of $d \neq d'$  is common for text explainers where the text input that a model sees maybe in form of \emph{n-grams} or \emph{Byte Pair Encoding (BPE)} \citep{sennrich-etal-2016-neural} vectors, whereas the explanation might be in an ``interpretable'' space such as presence/absence of words. For tabular data $d=d'$. We'll use the ``$'$" superscript to denote data in the explanation space, e.g., $X'_{orig}\in \mathbb{R}^{N_{orig} \times d'}$.
    
    \item Models \emph{Model v1} and \emph{Model v2} are denoted by the function $f$, parameterized by $\Psi_1$ and $\Psi_2$ respectively\footnote{As mentioned earlier, we discuss only one iteration of model improvement, but this discussion applies to the general case of learning $\Psi_{i+1}$, given $\Psi_{i}$.}. As examples, $\Psi$ may be coefficients in \emph{Logistic Regression} or weights in a \emph{Neural Network}.
     
    
    \item The explainer is denoted by the function $E(x;\theta, \Psi)$, where $x \in \mathbb{R}^d$ is an instance for which an explanation for its prediction by model $\Psi$ is sought. The explainer itself has parameters $\theta$, such as the number of features to be used in explanations \citep{lime}. 

    \item The explanation is a vector of weights $q \in \mathbb{R}^{d'}$ that explains the input $x$ in the explanation space, i.e., it applies to $x'\in \mathbb{R}^{d'}$. Intuitively, these weights indicate the importance of the corresponding feature.
    
    While this specific format for explanations is an assumption, it is common \citep{lime, shap, kim-etal-2020-interpretation, slack2021reliable} and allows our formulation to be broadly applicable.
    
    \item Finally, we account for two practical constraints in our setup:
    \begin{enumerate}
        \item $\mathcal{B}_E$, explanation budget: the number of instances whose explanations an ML practitioner might manually study.
        \item $\mathcal{B}_L$, labeling budget: the number of instances that annotators can label within one iteration. This is equivalent to the \emph{batch size} in AL.
    \end{enumerate}
    Typically, for real-world systems, $\mathcal{B}_E < \mathcal{B}_L < N_{inc}$. Some representative numbers are: $\mathcal{B}_E$ is in the order of hundreds, $\mathcal{B}_L$ is in the order of hundreds to thousands (depending on the labeling cost, e.g., skill required, number of annotators), and $N_{inc}$ may be arbitrarily large, potentially running into millions of instances.
    
\end{enumerate}

\subsection{Task Formulation}
\label{sec:task}
Given the above notation, we now revisit the workflow from Figure \ref{fig:workflow}:

\begin{enumerate}
    \item \emph{Step \texttt{C} in Figure \ref{fig:workflow}}: Explanations $E(x_i; \theta, \Psi_1)$ are sought for instances $x_i \in X_s$, where $X_s \subseteq X_{inc}$ is a set of instances randomly selected from $X_{inc}$, such that its size $N_s$ does not exceed the explanation budget $\mathcal{B}_E$. For each instance $x_i \in \mathbb{R}^d$, an explanation weight vector $q_i \in \mathbb{R}^{d'}$ is produced. 
    
    \item \emph{Step \texttt{D}, representing unintended model behavior}: Based on studying the explanations, the ML practitioner identifies instances in $X_s$ that indicate model behavior that is either unintended or in some sense, surprising. An example is that different labels are predicted for a pair of instances that are either similar or produce similar explanations. Intuitively, this might mean the model requires more such instances to confidently tell them apart. 

    Recall that the practitioner's goal is to select instances from $X_{inc}$ similar to the ones that participate in such pairs in $X_{s}$. And, since she can select only up to $\mathcal{B}_L$ instances, we want to favour instances in $X_s$ that participate in a large number of such pairs.  
    
    We represent this in the following manner (note that all $x$ appearing below belong to $X_s$):
    \begin{itemize}
        \item Let matrix $A \in \mathbb{R}^{N_s \times N_s}$ represent similarity between instances - either in terms of the vectors themselves, or their explanations. We combine them in the following way:
            \begin{equation}
                A_{ij} = (q_{i} \odot x_i')\cdot (q_{j}\odot x_j')^T    
            \end{equation}
            The ``$\odot$'' symbol represents the \emph{element-wise} product and the ``$\cdot$'' symbol denotes the \emph{dot product}.

        \item Let $B  \in \mathbb{R}^{N_s \times N_s}$ represent whether predicted labels are identical:
            \begin{align}
                B_{ij} = \begin{cases}
                            1  & f(x_i, \Psi_1) \neq f(x_j, \Psi_1) \\
                            0 & \text{ otherwise} 
                        \end{cases}
            \end{align}
            Note that pairs of instances are assigned a value of $1$ when the predicted labels are \emph{not} identical.

        \item Finally, given the identity vector $\mathbbm{1}_{N_s} \in \mathbb{R}^{N_s \times 1}$, we\footnote{The identity vector has a subscript that denotes its length. Also, we will assume, vectors are \emph{column} vectors.} define $C \in \mathbb{R}^{N_s}$:
        \begin{equation}
           C = (A \odot  B) \mathbbm{1}_{N_s}
        \end{equation}

        Consider the values for $A \odot  B$: 
        \begin{enumerate}
            \item +ve values indicate pairs of instances with different predictions, i.e., $B_{ij}=1$, but similar explanations, i.e., $A_{ij}> 0$ (preferred for retrieval).
            \item -ve values indicate different predictions, i.e., $B_{ij}=1$, but also different explanations, i.e., $A_{ij}<0$ (not preferred).
            \item $0$ values either indicate same predictions, i.e., $B_{ij}=0$ or different explanations , i.e., $A_{ij}=0$ (not preferred).
        \end{enumerate}
         $C_i$ provides a row-wise sum for instances $x_i$ in $A \odot  B$, quantifying the extent to which they are preferred during retrieval.

         As an examples, consider for $N_s=3$:
         \begin{align}
            &A= \begin{bmatrix} 1 & 0.7 & 0.8\\ 0.7 & 1 & 0.3\\0.8 & 0.3 & 1 \end{bmatrix}, 
            B= \begin{bmatrix} 0 & 1 & 1\\ 1 & 0 & 0\\1 & 0 & 0 \end{bmatrix}\\
            &A \odot  B = \begin{bmatrix} 0 & 0.7 & 0.8\\ 0.7 & 0 & 0\\0.8 & 0 & 0 \end{bmatrix}\\
            &C = (A \odot  B)\mathbbm{1}_{N_s} = \begin{bmatrix} 1.5\\0.7\\0.8\end{bmatrix}
         \end{align}
        Here, as per $B$,  $x_1$ and $x_2$ have the same predicted label, which is different from that of $x_0$. As per $A$, $x_0$ also has high explanation similarities to both $x_1$ and $x_2$. The desirability of $x_0$ is reflected in its high value in $C$. Intuitively, $C$ indicates preference weights for instances in $X_s$, to be used as queries for retrieval.
        
    \end{itemize}
    
    \item \emph{Step \texttt{D} (continued), retrieving instances from $X_{inc}$}: We now select $\mathcal{B_L}$ instances from $X_{inc}$ based on $C$. This is a human-in-the-loop activity, which we assume may be approximated with similarities via dot products. We detail this below.
    
    We compute the following score matrix $S\in \mathbb{R}^{N_{inc} \times N_{s}}$:
    \begin{equation}
        S=X_{inc}'(C\mathbb{I}_{d'}^T \odot X_s')^T
    \end{equation}
    Note that $X_{inc}'$ and $X_s'$ are representations in the explanation space. $C\mathbb{I}_{d'}^T \odot X_s'$ multiplies each vector in $X_s$ with the corresponding weight in $C$. Finally, $S$ computes the similarity, i.e., dot-product, between vectors in $X_{inc}$ and these weighted vectors from $X_s'$.

    To continue with our example, let's  consider the following $X_s'$ with $d'=2$:
    \begin{equation}
        X_s'= \begin{bmatrix} 0.7 & 0.2\\ 0.34 & 1.15 \\-0.1 & 3\end{bmatrix}, 
    \end{equation}
    Then, 
    \begin{align}
        C\mathbb{I}_{d'}^T = \begin{bmatrix} 1.5\\0.7\\0.8\end{bmatrix} \begin{bmatrix} 1 1 \end{bmatrix} = \begin{bmatrix} 1.5 & 1.5\\ 0.7 & 0.7 \\0.8 & 0.8\end{bmatrix}\\
        C\mathbb{I}_{d'}^T \odot X_s' = \begin{bmatrix} 1.5\times0.7 & 1.5\times 0.2\\ 0.7\times 0.34 & 0.7\times 1.15 \\0.8\times -0.1 & 0.8\times 3\end{bmatrix},  
    \end{align}
    
    $S$ is a symmetric matrix, where $S_{ij}$ denotes the similarity between $x_i' \in X_{inc}'$ and $x_j' \in X_s'$ (or indirectly, $x_i$ and $x_j$), accounting for the preferences encoded by $C$.

    To obtain the overall retrieval desirability for an instance in $X_{inc}'$, we compute its row-wise sum. We define the retrieval weights $W\in \mathbb{R}^{N_{inc}}$ as:
    \begin{equation}
        W = S \mathbbm{1}_{N_s}
    \end{equation}
    We select the top $\mathcal{B}_L$ instances from $X_{inc}$ based on $W$. We will refer to these instances as $X_{top} \in \mathbb{R}^{\mathcal{B}_L \times d}$.

    

    \item \emph{Steps \texttt{E} and \texttt{F}}: We obtain labels $Y_{top} \in \mathbb{R}^{\mathcal{B}_L}$ corresponding to $X_{top}$ via human annotation, and construct the following new dataset:
    \begin{align}
        &X_{new} =  \begin{bmatrix}
                        X_{orig} \\
                        X_{top}
                    \end{bmatrix}, X_{new} \in \mathbb{R}^{(N_{orig} + \mathcal{B}_L) \times d} \nonumber \\
        &Y_{new} = \begin{bmatrix}
                        Y_{orig} \\
                        Y_{top}
                    \end{bmatrix}, Y_{new} \in \mathbb{R}^{(N_{orig} + \mathcal{B}_L)}  \label{eqn:new_data}            
    \end{align}   
    $(X_{new}, Y_{new})$ is used to retrain $f$ obtaining new parameters $\Psi_2$.
  
\end{enumerate}

\textbf{These steps can be condensed into a single expression} - given matrices $A$ (this makes use of explanations) and $B$, the top-$\mathcal{B}_L$ instances from $X_{inc}$ based on these weights are picked:
\begin{equation}
\label{eqn:one_eqn}
    W = X_{inc}'(((A \odot  B) \mathbbm{1}_{N_s})\mathbb{I}_{d'}^T \odot X_s')^T \mathbbm{1}_{N_s}
\end{equation}

\subsection{Objective}
We want to minimize the generalization loss of $f$. In an AL setting the only labeled data available is $(X_{new}, Y_{new})$ (Equation \ref{eqn:new_data}) and therefore, a validation set must be sampled from it. To keep the notation simple, we use $\mathcal{L}_v$ to denote generalization loss, i.e., loss on a validation set, and our objective as:
\begin{equation}
\label{eqn:objective}
    \min_{\theta, \Psi} \mathcal{L}_v(X_{orig}, Y_{orig}, X_{inc}, \theta, \Psi)
\end{equation}

Note here that we optimize for the explanation parameters $\theta$ as well to refine them to be helpful in the data selection process, i.e., construction of $(X_{new}, Y_{new})$. 

\subsection{Metrics}
To measure competitiveness against standard AL approaches, we hold out a labeled test set $(X_{test}, Y_{test})$ that is large enough to reflect the true distribution. Such a dataset is not available in real-world AL setups, and is used here to measure the true accuracy of a model. We report model accuracy scores on this dataset, at various iterations of data being sampled from $X_{inc}$.

\begin{figure*}[ht]
\begin{center}
\centerline{\includegraphics[width=1.2\columnwidth]{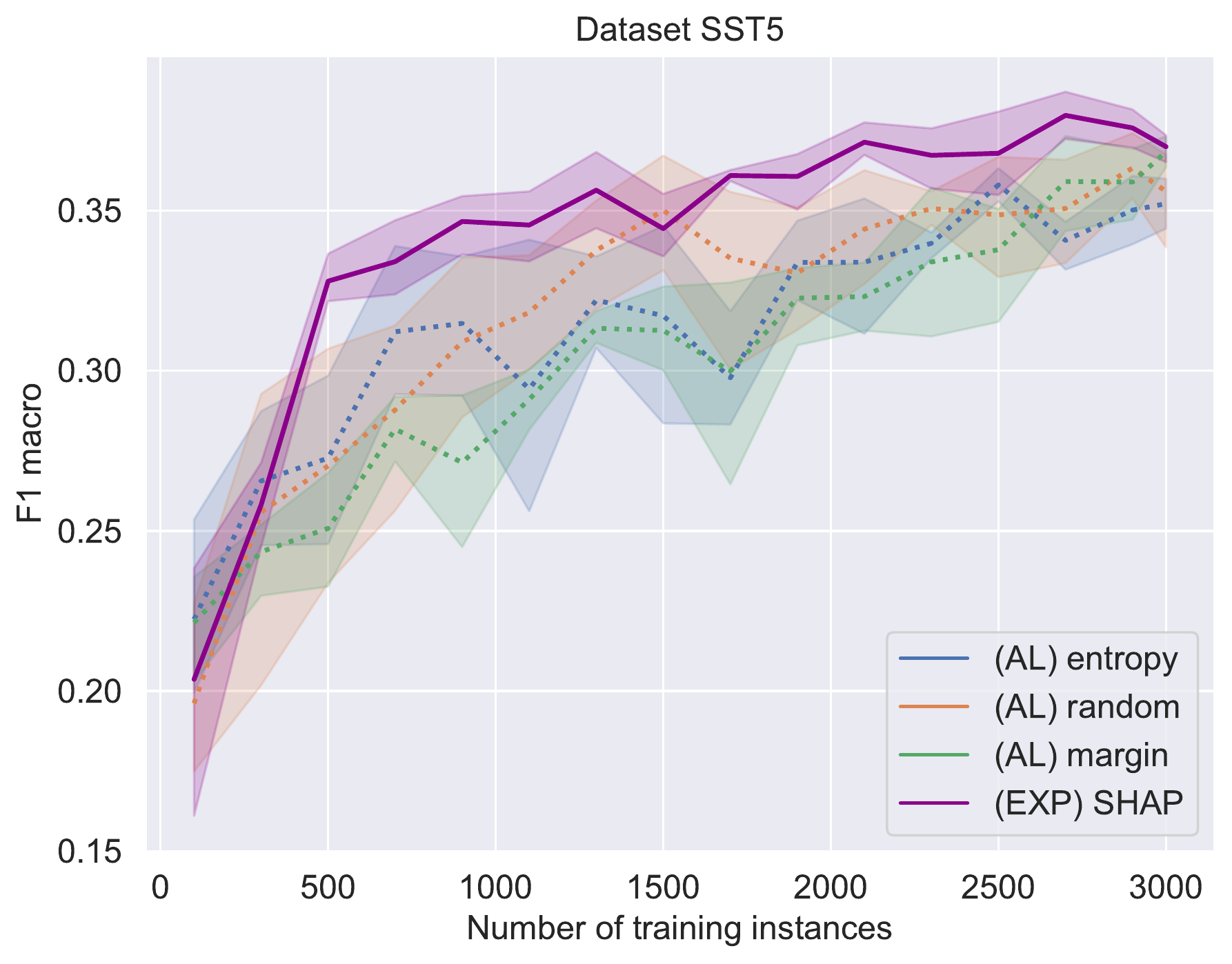}}
\caption{\emph{F1-macro} evaluated on $(X_{test}, y_{test})$ for the dataset SST5. The x-axis describes the number of instances used in each model training iteration. The solid line represents the accuracy for our explanation based AL (Section \ref{sec:task}). The dotted lines represent results for different popular AL methods: \emph{entropy sampling}, \emph{random sampling}, \emph{maximum margin sampling}. The band around each line indicates 95\% confidence intervals across 4 runs.}
\label{fig:plot}
\end{center}
\end{figure*}

\subsection{Extensions}
While we looked at at one form of unintended behavior here - different predictions but similar explanations - it is possible to define others. These may be defined based on $(X_{orig},Y_{orig})$ as well. Some examples are:
\begin{enumerate}
    \item For an instance $x_i \in X_{orig}$, the predicted label is incorrect, but the explanations for both predicted label and true label are similar. It is possible to specify this behavior only if the explainer can generate explanations for membership to any class, e.g., LIME, SHAP. 
    
    Intuition: the model is misaligned with respect to its mapping from features to labels.
    \item For a pair of instances with different true labels, the explanations for membership to these labels are similar. 
    
    Intuition: the model is unable to strongly discriminate between the two classes for these instances, and might need more similar instances to learn.
\end{enumerate}
These criteria can be easily included in our formulation by appropriately  defining matrices $A$ and $B$.

\subsection{Review of Assumptions}
In our formulation we make two assumptions:
\begin{itemize}
    \item The format of explanations as a weight vector. As mentioned earlier, this is indeed a common format, and allows the formulation to be broadly applicable.
    \item Approximating the human-in-the-loop process with retrieval based on dot-product based similarities. While this is probably reasonable, we require user studies to validate its adequacy. 
\end{itemize}
\qedsymbol{} 

We note that the above formulation is generic and applicable to different kinds of data, e.g., text, images, tabular, different explainers, as well as different models.

\section{Experiments}

We have begun empirical comparisons to standard AL techniques. While our goal is to cover a diverse set of data, models and explainers, we present initial results on the dataset SST-5 \citep{socher-etal-2013-recursive}  using SHAP \citep{shap}, specifically Partition SHAP, as the explainer and \emph{Support Vector Machine} with linear kernel as our model (the \emph{scikit-learn} \citep{sklearn_api} library is used). We use the \emph{F1-macro} score to report accuracy; this metric is used since it accounts for class-wise accuracies even when there is class imbalance. The optimization in Equation \ref{eqn:objective}  is solved using \emph{Bayesian Optimization} \citep{bayesopt}, since they enable us to minimize non-differentiable functions. This is an important consideration since we are not guaranteed differentiability, e.g., when using a Decision Tree as our model. We specifically use the \emph{Ray Tune} \citep{liaw2018tune} and \emph{Optuna} \citep{optuna_2019} libraries. For AL, we use the \emph{modAL} library \citep{modAL2018}.

For reasons of tractability, the joint optimization over $\theta$ and $\Psi$ (Equation \ref{eqn:objective}) is decomposed into the following nested optimization:
\begin{enumerate}
    \item The search space of the explanation parameters $\theta$ is explored by the Bayesian Optimizer. The SHAP parameters we varied are maximum number of predictions which model $f$ makes for explaining one instance and maximum number of predictions in one model invocation. 
    \item The model selection search space is explored by standard cross-validation with grid-search over hyperparameters: in this case, the regularization coefficient $C$.
\end{enumerate}

Other relevant details:
\begin{enumerate}
    \item The text representation used is \emph{Universal Sentence Encoding (USE)} \citep{USE}.
    \item Experiment settings:
    \begin{itemize}
        \item $N_{orig}=100$, $N_{inc}=2900$, $N_{test}=2000$.
        \item Labeling budget (or batch size in AL), $\mathcal{B}_L=200$, explanation budget $\mathcal{B}_L = 200$.
    \end{itemize}
    \item Comparisons against:
    \begin{itemize}
        \item AL query strategies: entropy-based sampling\footnote{This selects instances with high entropy values over prediction probabilities across classes.}, margin-based sampling\footnote{Selects instances that have a small difference between the prediction probabilities of the two most confident classes.} \citep{margin10.1007/3-540-44816-0_31}. See \citet{settles2009active} for an overview.
        \item Baseline: we use a \emph{random} strategy, which selects $\mathcal{B}_L$ instances from $X_{inc}$ uniformly at random with no replacement.
    \end{itemize}
\end{enumerate}

We visualize our results in Figure \ref{fig:plot}. ``AL'' or ``EXP'' in the legend denote whether a strategy comes from standard AL or is based on an explainer.  We observe that the explanation based query strategy performs better than other standard AL techniques. It achieves higher scores right at the first few iterations and reaches a plateau in performance. We also note that the AL query strategies are not significantly better than random selection.

\section{Conclusions and Future Work}
\label{sec:conclusion}
In this short paper, we investigated a specific use of XAI: using it to select model training data. We showed that this is equivalent to performing Active Learning, with a human-in-the-loop as part of the query strategy. Further, we mathematically approximated this workflow, so that it may be conveniently studied and empirically compared to other Active Learning techniques.  We presented some initial results; these look promising, and we hope to continue the empirical analyses to definitively establish the utility of XAI in this setup. 

Our future work would focus on: (a) validating our approximation for the human-in-the-loop process via user-studies, (b) broadening the scope of this study by using different classifiers, text representations, datasets and AL techniques, especially those recently proposed, e.g., \citet{zhdanov2019diverse}, \citet{CARDOSO2017313}, and (c) exploiting the differentiability of our formulation (Equation \ref{eqn:one_eqn}) to \emph{learn} an AL strategy.

\bibliography{refs}
\bibliographystyle{icml2023}

\newpage
\appendix
\onecolumn


\end{document}